\documentclass{article}

\usepackage{PRIMEarxiv}

\usepackage[utf8]{inputenc} 
\usepackage[T1]{fontenc}    
\usepackage{hyperref}       
\usepackage{url}            
\usepackage{booktabs}       
\usepackage{amsfonts}       
\usepackage{nicefrac}       
\usepackage{microtype}      
\usepackage{lipsum}
\usepackage{fancyhdr}       
\usepackage{graphicx}       
\usepackage{adjustbox}
\usepackage{multirow}
\usepackage{enumitem}  
\usepackage{amsmath}
\graphicspath{{media/}}     

\title{MedDINOv3: How to adapt vision foundation models for medical image segmentation?
}

\author{
  Yuheng Li$^{1}$, 
  Yizhou Wu$^{2}$, 
  Yuxiang Lai$^{3}$, 
  Mingzhe Hu$^{3}$, 
  Xiaofeng Yang$^{1,3,4,*}$ \\
  \\
  $^{1}$Department of Biomedical Engineering, Georgia Institute of Technology, Atlanta \\
  $^{2}$Department of Electrical and Computer Engineering, Georgia Institute of Technology, Atlanta \\
  $^{3}$Department of Computer Science, Emory University, Atlanta \\
  $^{4}$Department of Radiation Oncology, Emory University School of Medicine, Atlanta \\
  $^{*}$Email: xiaofeng.yang@emory.edu
}

\begin{document}
\maketitle

\begin{abstract}

Accurate segmentation of organs and tumors in CT and MRI scans is essential for diagnosis, treatment planning, and disease monitoring. While deep learning has advanced automated segmentation, most models remain task-specific, lacking generalizability across modalities and institutions. Vision foundation models (FMs) pretrained on billion-scale natural images offer powerful and transferable representations. However, adapting them to medical imaging faces two key challenges: (1) the ViT backbone of most foundation models still underperform specialized CNNs on medical image segmentation, and (2) the large domain gap between natural and medical images limits transferability. We introduce \textbf{MedDINOv3}, a simple and effective framework for adapting DINOv3 to medical segmentation. We first revisit plain ViTs and design a simple and effective architecture with multi-scale token aggregation. Then, we perform domain-adaptive pretraining on \textbf{CT-3M}, a curated collection of 3.87M axial CT slices, using a multi-stage DINOv3 recipe to learn robust dense features. MedDINOv3 matches or exceeds state-of-the-art performance across four segmentation benchmarks, demonstrating the potential of vision foundation models as unified backbones for medical image segmentation. The code is available at https://github.com/ricklisz/MedDINOv3. 
\end{abstract}

\keywords{Self-supervised learning \and Foundation model \and Medical image segmentation}

\section{Introduction}

Medical imaging modalities such as computed tomography (CT) and magnetic resonance imaging (MRI) are central to modern radiology, enabling detailed visualization of anatomical structures and abnormalities. Accurate segmentation of organs-at-risk (OARs) and tumors is crucial for treatment planning and disease monitoring \cite{ji2022amos, li2025automatic}, yet manual annotation is labor-intensive and time-consuming \cite{tang2022self}. Deep learning has shown great promise in automating this process; however, most existing approaches rely on highly specialized architectures trained for individual datasets or organ systems \cite{isensee2021nnu, hatamizadeh2022swinunetr}, limiting their generalization across modalities and institutions \cite{ma2023medsam}.

Foundation models (FMs) offer a promising solution as unified visual backbones, pretrained on large-scale unlabeled data and adaptable across diverse downstream tasks \cite{bommasani2021opportunities, zhou2023comprehensive, li2024polyp, wang2025dinov3}. Self-supervised learning enables training directly from raw pixels without manual annotations, producing transferable representations \cite{caron2021emerging, radford2021clip, tschannen2025siglip2}. Recent advances such as DINOv2 \cite{oquab2023dinov2} and DINOv3 \cite{simeoni2025dinov3} have demonstrated remarkable success in natural images, producing strong global and local features for classification, detection, and segmentation tasks.

However, due to privacy concerns, it is infeasible to obtain billion-scale data for training medical vision foundation models from scratch. This raises a natural question: \textit{can representations learned from web-scale natural images be effectively transferred to radiological imaging?} Our empirical results suggest that DINOv3 provides promising performance for medical image segmentation. Nevertheless, we identify two key challenges in adapting vision foundation models to CT and MRI segmentation: 1). current FM backbones are based on Vision Transformers (ViTs), which still lag behind strong CNN baselines in dense prediction tasks \cite{isensee2024nnu}; 2). a substantial domain gap between natural and medical images prevents direct transfer of pretrained representations \cite{baharoon2024dinov2}.

\begin{figure}[h!]
    \centering
    \includegraphics[width=1.0\textwidth]{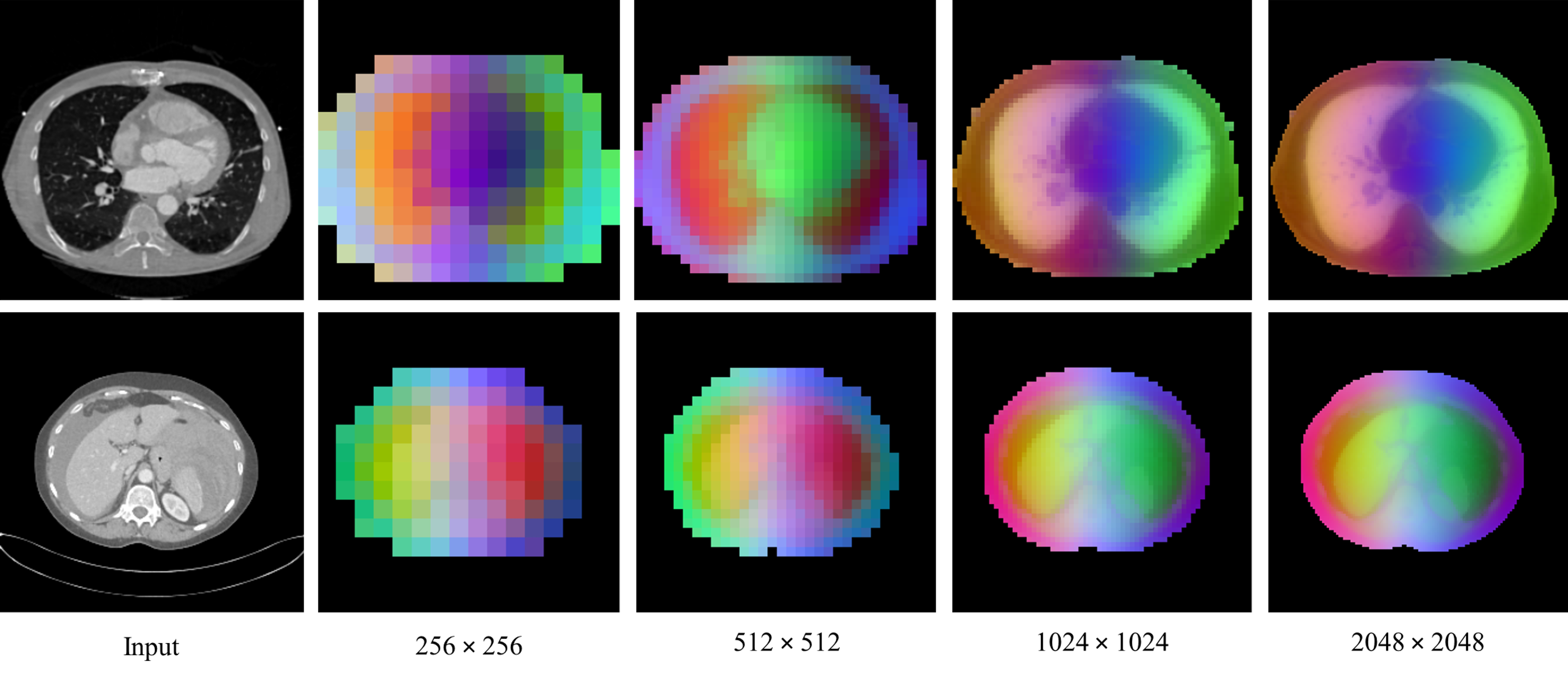}
    \caption{MedDINOv3 PCA maps at progressively higher resolution. We visualize dense features of MedDINOv3 by mapping the first three components of a PCA computed over the feature space to RGB. We mask the feature maps to focus on the CT foreground. }
    \label{fig:pca}
\end{figure}

In this work, we propose MedDINOv3, an effective framework to adapt vision foundation model for medical image segmentation. First, we revisit plain ViTs for 2D medical image segmentation. While ViT has proven to be scalable for large-scale pretraining \cite{zhai2022scaling}, it still requires tailored components such as ViT-Adapter \cite{zhang2023vitadapter} or Mask2Former \cite{cheng2022mask2former} to achieve good segmentation performances. In medical imaging, existing transformer-based designs often fall back on heavy convolutional components and still underperform strong CNN baselines \cite{hatamizadeh2022swinunetr, hatamizadeh2022unetr, xie2021cotr, xie2022unimiss, chen2021transunet, zhou2021nnformer}. Inspired by recent works that deconstruct transformer architectures for segmentation \cite{wald2025primus, kerssies2025eomt}, we propose a simple and effective transformer architecture for 2D medical image segmentation. MedDINOv3 leverages the DINO ViT as a strong vision encoder and introduces multi-scale feature aggregation by reusing patch tokens from intermediate transformer blocks. This hierarchical representation provides richer spatial contexts to the decoder, mitigating the weak locality bias of ViTs. Second, we perform domain-adaptive pretraining of MedDINOv3 on CT-3M, a large-scale collection of CT images, to better align the model with radiological data distributions. We found that gram anchoring, a mechanism to prevent local feature from collapsing, is optional to our pretraining framework. We systematically examine the three-stage pretraining recipe of DINOv3 and quantify the contribution of each stage to segmentation performance. After adapting to CT domain, our pretrained MedDINOv3 produces smooth feature maps at consistently higher resolutions (Fig.\ref{fig:pca}). 

We summarize our main contributions as follows:
\begin{itemize}[leftmargin=*, itemsep=0.3em]
    \item \textbf{A simple ViT architecture for 2D medical segmentation.}  
    We revisit plain Vision Transformers and propose an effective design for 2D medical image segmentation. Two key refinements—(i) \emph{multi-scale token aggregation} from intermediate patch tokens and (ii) \emph{high-resolution training}—raise ViT-B performance on AMOS22 from 78.39\% to 85.51\% DSC.  

    \item \textbf{Domain-adaptive pretraining on CT-3M.}  
    We curate \emph{CT-3M}, a large-scale collection of axial CT slices from 16 datasets, and adapt DINOv3 via a three-stage process: (1) global/local self-distillation (DINOv2-style), (2) gram anchoring to stabilize patch-level consistency, and (3) high-resolution adaptation. We systematically examine each stage and quantify its impact on downstream segmentation.  

    \item \textbf{State-of-the-art results across four public CT/MRI benchmarks.}  
    On four diverse benchmarks (AMOS22, BTCV, KiTS23, LiTS), MedDINOv3 outperforms or matches strong CNN and transformer baselines. It surpasses nnU-Net on OAR segmentation (+2.57\% DSC on AMOS22 and +5.49\% DSC on BTCV), while achieving comparable tumor segmentation on KiTS23 (70.68\% DSC) and LiTS (75.28\% DSC). These results highlight the effectiveness of domain-adaptive pretraining for transferring vision foundation models to radiology.
\end{itemize}

\section{Related work}
\subsection{Medical vision foundation models}
Self-supervised learning has emerged as a key strategy for developing medical vision foundation models, motivated by the scarcity of annotated medical data. Models Genesis \cite{zhou2021models} demonstrated the benefits of pretext reconstruction tasks on CT and MRI, while more recent frameworks such as SwinUNETR with SSL pretraining \cite{tang2022self} showed substantial improvements on 3D CT benchmarks. Masked image modeling (MIM) has also been applied in the medical domain, with studies showing that MIM-pretrained encoders significantly improve performance on organ segmentation datasets \cite{li2025automatic, wald2025revisiting, li2024anatomask}. Parallel work has explored adapting natural-image FMs such as DINOv2 to radiology: Baharoon et al.~\cite{baharoon2024dinov2} demonstrated that natural-image SSL features transfer well to classification, though performance in segmentation lags behind domain-pretrained models. Our work bridges this performance gap by performing domain-adaptive SSL pretraining at scale.

\subsection{Vision transformers in medical image segmentation}
Vision Transformers have been widely explored in medical image segmentation. TransUNet \cite{chen2021transunet} incorporated transformer layers into the bottleneck of a U-Net. LeViT-UNet \cite{xu2023levit} followed a similar design with efficient attention. UTNet \cite{gao2021utnet} employed Transformer blocks at multiple resolutions, and CoTr \cite{xie2021cotr} leveraged a single Transformer to jointly model features across resolutions. Among the most influential works, UNETR \cite{hatamizadeh2022unetr} directly employed a ViT encoder, representing a shift toward transformer-heavy designs. Following this, SwinUNETR \cite{hatamizadeh2022swinunetr} integrated swin transformers. Most recent works such as Primus \cite{wald2025primus} deconstruct the complex decoders and show that an encoder-only architecture can approach CNN performance. Despite this progress, recent benchmark studies emphasize that CNNs like nnU-Net remain strong baselines \cite{isensee2024nnu}. Our work aims to fully realize the potential of ViTs in medical segmentation by refining the architecture design and performing domain-adaptive SSL pretraining. 

\begin{figure}[h!]
    \centering
    \includegraphics[width=1.0\textwidth]{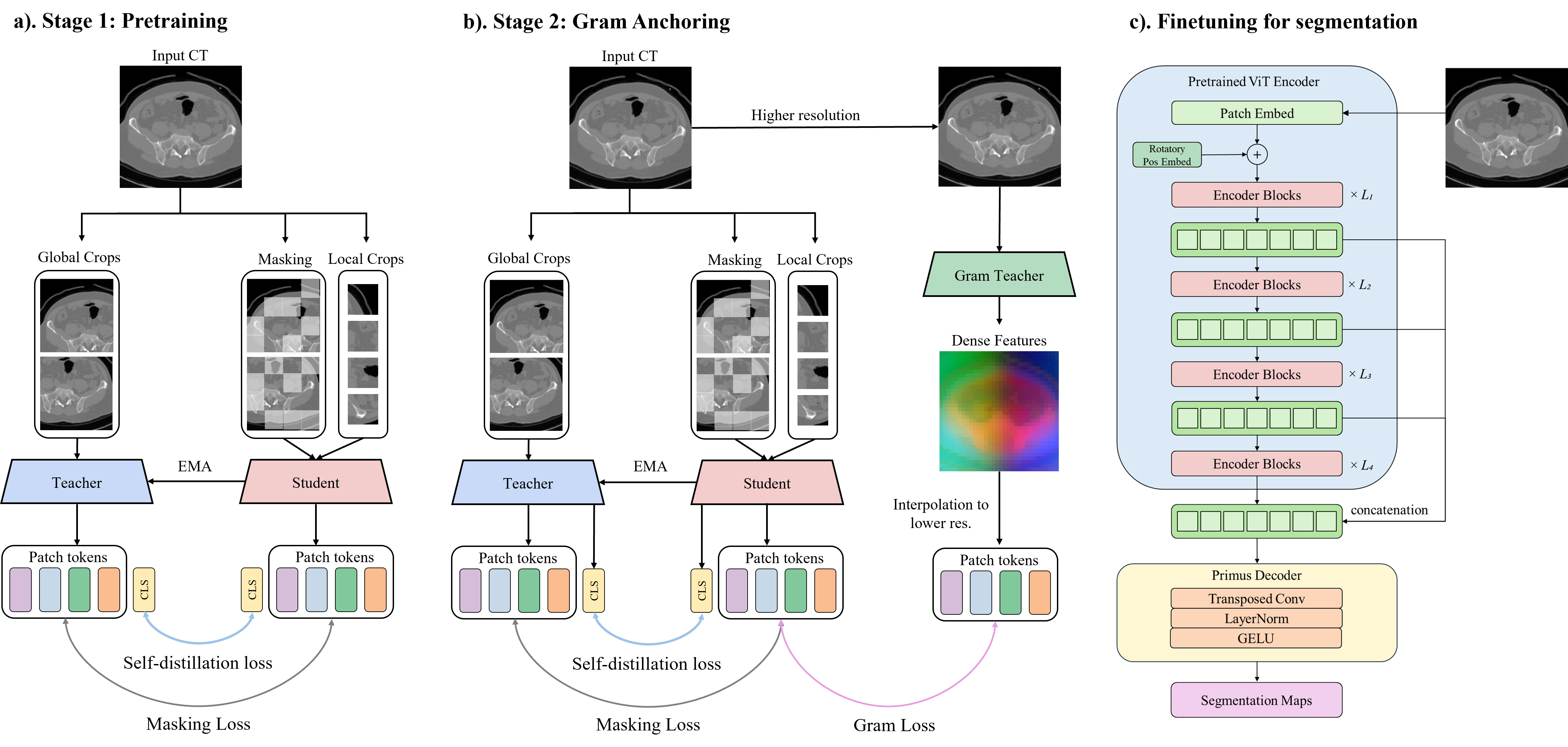}
    \caption{Overall framework of MedDINOv3. a). Stage 1: Given an input CT, we feed the global crops to the teacher model, local and masked crops to the student. Self-distillation loss is applied to the CLS tokens and masking loss applied to dense patch tokens.  b). Stage 2: Adds gram anchoring. Gram teacher sees a higher resolution global crop and outputs dense feature maps, resized to match student resolution. Stage 3: Both student and teacher are trained with higher-res CT inputs (not shown). c). Finetuning pretrained MedDINOv3 for segmentation with proposed architecture. }
    \label{fig:pipeline}
\end{figure}

\section{Method}
While vision foundation models pretrained on large datasets have demonstrated remarkable performance in natural images, directly adapting them for medical imaging remains non-trivial, due to considerable gaps in representations and different architectural designs. First, we propose iterative refinements on vision transformer for medical image segmentation, enabling integration of pretrained SSL models. Next, we perform domain-adaptive pretraining on CT images using DINOv3, a state-of-the-art SSL method known for learning superior dense features. We then transfer the learned representations to various medical segmentation tasks. Our framework is shown in Figure \ref{fig:pipeline}.

\subsection{Rethinking transformers for medical image segmentation}
Despite promising scalability in natural images, vision transformers have not yet achieved consistent gains for medical image segmentation. Prior studies show that transformer blocks in segmentation models contribute very little to final performance \cite{wald2025primus} and still underperform strong CNN baselines \cite{isensee2021nnu}. To enforce attention usage, Primus proposes a simple transformer architecture that uses a lightweight transposed convolution decoder, achieving satisfactory performances in 3D volumetric segmentation. However, Primus still lacks the spatial priors essential for dense segmentation. Furthermore, Primus encoder modifies its patch size from 16 to 8, which remains compute heavy and does not suit DINOv3-style pretraining. Motivated by these limitations, we revisit plain ViTs for 2D medical image segmentation and propose a stepwise refinement to adapt DINOv3 backbone into an effective segmentation backbone (Table~\ref{tab:ablation_arch}).

\paragraph{Development datasets}
We select AMOS22 to train and evaluate each refinement step. AMOS22 \cite{ji2022amos} is an abdominal organ segmentation dataset of 300 CT volumes and 60 MRI volumes with 15 annotated organs. We follow a similar strategy as Primus \cite{wald2025primus} by training and evaluating with only one fold (80/20 split) of the default five-fold cross-validation scheme. We train for a total of 1000 epochs following default nn-UNet settings, with hyperparameters adapted from Primus.

\paragraph{Baseline}
We form our baseline using DINOv3 ViT encoder (ViT-B) and Primus decoder composed of back-to-back transposed convolution, LayerNorm, and GELU activation, to upsample patch tokens into the full-resolution segmentation map. This design minimizes convolutional influence and maximizes the impact of transformer's representations. Starting from random initialization, this configuration provides a reasonable baseline, but lags behind supervised CNNs.

\paragraph{Leveraging pretrained DINOv3}
To improve representation quality, we initialize the ViT encoder with DINOv3 pretrained on LVD-1689M. We observe considerable boost in DSC by 2.96\% over random initialization, highlighting the transferability of web-scale SSL features to medical segmentation. 

\paragraph{Multi-scale token aggregation}
Observing that Primus only uses the last transformer block as input to the decoder, we hypothesize that this lack of hierarchical priors prevents ViTs from learning strong local features. We propose to reuse patch tokens from multiple intermediate layers (blocks 2, 5, 8, 11) and concatenate them as input to the decoder. This step enriches the spatial priors that are otherwise weak in ViTs. As shown in Table~\ref{tab:ablation_arch}, incorporating multi-scale features considerably improved DSC by 2.10\% in AMOS22.

\paragraph{Higher resolution training}
To preserve local information, Primus propose to decrease the patch size during tokenization from 16 to 8, and found this beneficial for 3D volumetric segmentation. However, existing vision FMs are rarely pretrained using patch size of 8, possibly due to increased computational overheads. As an alternative, we propose to conduct high resolution segmentation training by resampling axial slices to thinner spacing. Following DINOv3 \cite{simeoni2025dinov3}, we maintain an input resolution of $896\times 896$. As shown in Table~\ref{tab:ablation_arch}, increasing resolution from $640\times 640$ to $896\times 896$ improved DSC by 2.06\% on AMOS22. 

\begin{table}[t]
  \centering
  \caption{Ablation study on refinements for adapting ViT backbone for segmentation on AMOS22.}
  \label{tab:ablation_arch}
  \begin{tabular}{cccc|c}
    \toprule
    Encoder init. & Decoder & Multi-scale features & Resolution & DSC (\%) \\
    \midrule
    randinit. & Primus & $\times$ & 640 $\times$ 640  & 78.39 \\
    DINOv3 & Primus & $\times$ & 640 $\times$ 640 & 81.35 \\
    DINOv3 & Primus & $\checkmark$ & 640 $\times$ 640 & 83.45 \\
    DINOv3 & Primus & $\checkmark$ & 896 $\times$ 896 & 85.51 \\
    \bottomrule
  \end{tabular}
\end{table}

\subsection{Domain-adaptive pretraining on CT-3M}
With an effective architecture, we aim to pretrain MedDINOv3 using a diverse, large-scale medical imaging dataset CT-3M, to better align its representations to medical imaging. We follow the 3-stage pretraining recipe developed by DINOv3 \cite{simeoni2025dinov3}.

\paragraph{Data curation}
We curated a large-scale CT dataset CT-3M totaling 3,868,833 axial slices, aggregated from 16 publicly available datasets. Specifically, our datasets include: BTCV \cite{landman2015miccai}, Pancreas-CT (TCIA) \cite{roth2015deeporgan}, CHAOS \cite{kavur2021chaos}, LiTS \cite{bilic2023liver}, KiTS \cite{heller2023kits21}, WORD \cite{luo2022word}, AbdomenCT-1K \cite{ma2021abdomenct1k}, AMOS22 \cite{ji2022amos}, and five CT tasks from Medical Segmentation Decathlon (Liver, Lung, Pancreas, Hepatic Vessel, Spleen, Colon) \cite{antonelli2022medical}, CT-ORG \cite{ctorg2020}, TotalSegmentator \cite{wasserthal2023totalsegmentator} and AbdomenAtlas 3.0 \cite{bassi2025radgpt}. This data curation provides broad anatomical coverage (over 100 structures) across abdominal, thoracic, and pelvic regions, ensuring both scale and heterogeneity for domain-adaptive pretraining. All 3D volumes were resampled to an in-plane spacing of 0.45 mm and 0.45 mm, and then resized to uniform resolution 256 $ \times $ 256. 

\paragraph{Stage 1}
We pretrain using the original DINOv2 losses: an image-level objective $L_\text{DINO}$ enforcing global-local crop invariance, a patch-level latent reconstruction objective $L_\text{iBOT}$ which learns the local patch correspondence, and a regularization loss $L_\text{Koleo}$ encouraging the features within a batch to spread uniformly in the latent space. The stage 1 loss is defined as follows:

\begin{equation}
\mathcal{L}_{\text{Stage1}} = \mathcal{L}_{\text{DINO}} + \mathcal{L}_{\text{iBOT}} + 0.1 \cdot \mathcal{L}_{\text{Koleo}}.
\end{equation}

\paragraph{Stage 2} 
However, DINOv2 training showed that global losses tend to dominate as training progresses, leading to a slow erosion of patch-level quality \cite{simeoni2025dinov3}. To address this, stage 2 introduces gram anchoring to mitigate the degradation of patch-level consistency. The motivation is that global and local objectives are only weakly correlated, while optimizing for global consistency often harms local feature quality. Gram anchoring regularizes Gram matrix, the matrix of all pairwise dot products of patch features in an image. Specifically, we encourage the Gram matrix of the student to align with that of an earlier model, referred to as the Gram teacher. The Gram teacher is chosen from an early checkpoint of the EMA student network, which retains stronger dense features. Formally, given an image with $P$ patches, and a network that operates in dimension $d$. Let $\mathbf{X}_S$ and $\mathbf{X}_G$ denote the $P \times d$ matrix of $L_2$-normalized local features of the student and the Gram teacher respectively. We define the loss $\mathcal{L}_\text{Gram}$ as follows:

\begin{equation}
\mathcal{L}_\text{Gram} = 
\left\lVert \mathbf{X}_S \cdot \mathbf{X}_S^\top 
- \mathbf{X}_G \cdot \mathbf{X}_G^\top \right\rVert_F^2.
\end{equation}

This loss is only computed on the global crops across all patch tokens. In our implementation, we start this stage after 100k iterations of pretraining, for a total of 20k iterations. We also leverage higher-resolution CT images as the input to the Gram teacher. Specifically, we feed images at 512 $\times$ 512 into the Gram teacher, then downsample the resulting feature maps by a factor of 2 to match the spatial dimensions of the student output. The stage 2 loss is defined as:

\begin{equation}
\mathcal{L}_\text{Stage2} = 
\mathcal{L}_\text{DINO} + \mathcal{L}_\text{iBOT} 
+ \mathcal{L}_\text{Koleo} 
+ 2* \mathcal{L}_\text{Gram}.
\end{equation}

\begin{figure}[h!]
    \centering
    \includegraphics[width=1.0\textwidth]{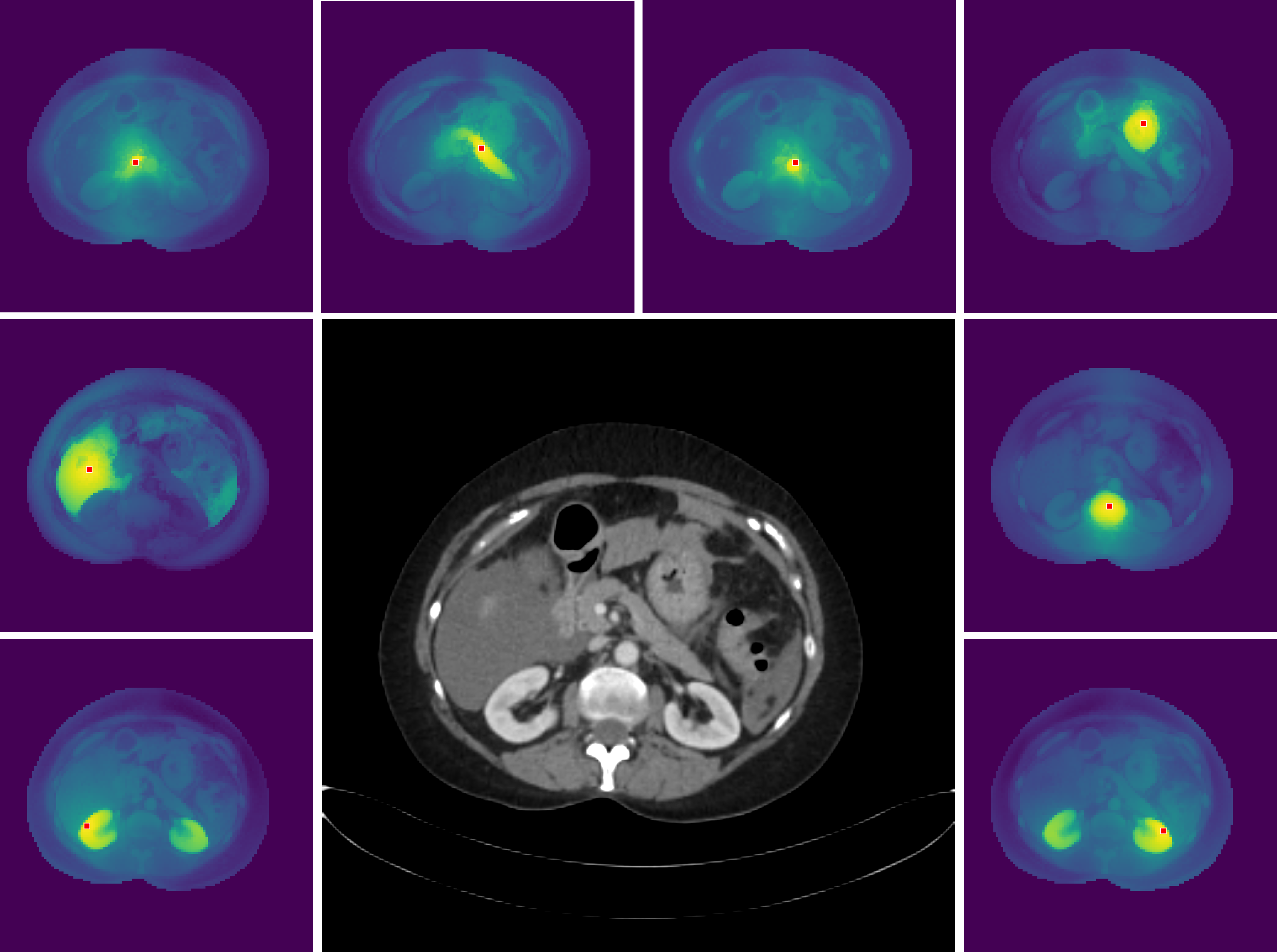}
    \caption{High-resolution dense features of MedDINOv3. We visualize the cosine similarity maps between the patches marked with a red dot and all other patches. Input image at 2048 $\times$ 2048.}
    \label{fig:dense_feature}
\end{figure}

\paragraph{Stage 3}  
The final stage adapts the pretrained model to process higher-resolution images, which is particularly relevant to our tasks. We follow DINOv3 by mixing global and local crops of various resolutions (e.g., global crops 512–768, local crops 112–336). Importantly, we retain gram anchoring to ensure that patch similarity structures remain stable. This stage lasts for 10k iterations. We found that high-resolution adaptation substantially improves the model's dense feature (Fig. \ref{fig:dense_feature}).  

\paragraph{Implementation details}  
We pretrain MedDINOv3 for a total of 120k iterations, using a global batch size of 512. We initialize with the DINOv3 ViT-B checkpoint pretrained on LVD-1689M. For stage 1, we use a learning rate of 2e-4 and train for 100k iterations. For stage 2, we select a EMA model at 20k iteration as our Gram teacher, and continue pretraining with gram anchoring objective for 10k iterations. This stage uses a learning rate of 5e-5. For step 3, we initialize teacher and student with the previous model from stage 2, and train for another 10k iterations using a learning rate of 2.5e-5. 

\section{Results}
\subsection{Experiment settings}
\paragraph{Evaluation Dataset} 
To comprehensively evaluate existing 2D segmentation methods, we conducted extensive experiments on four publicly available datasets. These datasets cover a wide spectrum of tasks in medical image segmentation (i.e. OARs and tumor), spanning diverse imaging modalities (e.g., CT and MRI). In addition to AMOS22, we added the following datasets: 1). KiTS23 \cite{heller2023kits21}, kidney tumor dataset with 489 CT volumes with annotations provided for kidney, tumor and cysts; 2). LiTS \cite{bilic2023liver}, liver tumor segmentation dataset with 131 CT volumes with annotated liver and tumor classes; 3). BTCV \cite{landman2015miccai}, a contrast-enhanced abdominal CT dataset with 50 scans with manual segmentation of 13 organs. Due to high computational costs, we trained and evaluated with only one fold (80/20 split) of the default five-fold  cross-validation scheme. For evaluation metrics, we use the standard dice similarity coefficient (DSC) and normalized surface dice (NSD). 

\paragraph{Implementation details} 
We performed training and evaluation within the nnU-Net framework in PyTorch. We summarize the preprocessing details used for each dataset in Table. All models were trained for 1,000 epochs, each consisting of 250 steps. Unless stated otherwise, input patch size, batch size, and voxel spacing follow the specific configurations defined by the respective nnU-Net plans. We report the following methods for comparison and detail the training configurations:
\noindent
\begin{enumerate}[leftmargin=*, labelsep=0.5em]
    \item \textbf{nnU-Net} \cite{isensee2021nnu}: Strongest supervised CNN baseline. Following the default nnU-Net v2 configuration, we use a learning rate of $1 \times 10^{-2}$, weight decay of $3 \times 10^{-5}$, gradient clipping set to 12, and the SGD optimizer with Nesterov momentum (0.99), along with the default nnU-Net PolyLR scheduler.

    \item \textbf{SegFormer} \cite{xie2021segformer}: A hierarchical transformer model with a lightweight MLP decoder architecture to directly fuse multi-level features. We adjust the learning rate to $5 \times 10^{-5}$ and use the AdamW optimizer.
    
    \item \textbf{DINO U-Net } \cite{gao2025dino}: A newly developed U-Net architecture supporting DINOv3 integration. We use the standard nnU-Net hyperparameters, but with the encoder backbone frozen, as described in the original paper.

    \item \textbf{MedDINOv3}: Our proposed method. We adapt the hyperparameters from Primus and use a higher input resolution of $896 \times 896$. The Primus model is trained using a learning rate of $3 \times 10^{-4}$ and a weight decay of $5 \times 10^{-2}$. We apply a DropPath rate of 0.2 and use LayerScale with a value of $1 \times 10^{-5}$. The optimizer is AdamW, configured with betas set to (0.9, 0.98).
\end{enumerate}


\subsection{Comparisons with state-of-the-art methods}
We compare MedDINOv3 against CNN and transformer baselines on four public segmentation benchmarks (Table~\ref{tab:comparison_sota}). Our MedDINOv3 consistently outperforms the best baseline nnU-Net in AMOS22 by 2.6\% DSC and BTCV by 5.49\% DSC. On tumor segmentation datasets, MedDINOv3 reaches 70.68 DSC on KiTS23 and 75.28 DSC on LiTS, performing on par with nnU-Net. In terms of boundary accuracy, MedDINOv3 maintains strong NSD scores for organ-at-risk segmentation, while only slightly trailing nnU-Net on tumor datasets. DINO U-Net, despite leveraging the DINOv3 foundation model, did not outperform nn-UNet, possibly due to its reliance on hierarchical CNN decoders. SegFormer, developed for natural image segmentation, underperforms across all datasets, reflecting its weaker inductive bias and reliance on large-scale labeled data. Overall, these results demonstrate that combining architectural improvements and domain-adaptive pretraining produces transferable representations for medical imaging, narrowing the gap with and in several cases surpassing the long-established nnU-Net baseline.

\begin{table}[htbp]
\centering
\caption{Performances on four public segmentation benchmarks. We report average DSC and NSD of all datasets. Due to
 computational constraints, the results are only calculated for one fold of a 5-fold cross-validation.}
 \label{tab:comparison_sota}
\begin{adjustbox}{max width=\textwidth}
\begin{tabular}{lcccc|cccc}
\toprule
\multirow{2}{*}{Method} & \multicolumn{4}{c}{{DSC (\%)}$\uparrow$} & \multicolumn{4}{c}{{NSD (\%)}$\uparrow$} \\
\cmidrule(lr){2-5} \cmidrule(lr){6-9}
 & AMOS22 & KiTS23 & LiTS & BTCV & AMOS22 & KiTS23 & LiTS & BTCV \\
\midrule
nnU-Net     & 84.81 & 69.15 & 75.00 & 73.30 & 73.98 & \textbf{64.85} & \textbf{53.02} & 64.66 \\
SegFormer   & 78.50 & 57.73 & 65.45 & 37.04 & 65.20 & 47.65 & 35.98 & 22.39 \\
Dino U-Net (B)   & 80.90 & 59.77 & 72.89 & 66.88 & 67.00 & 51.05 & 48.25 & 56.23 \\
\textbf{MedDINOv3 (ours)}  & \textbf{87.38} & \textbf{70.68} & \textbf{75.28} & \textbf{78.79} & \textbf{77.15} & 62.67 & 53.01 & \textbf{70.38} \\
\bottomrule
\end{tabular}
\end{adjustbox}
\end{table}

\begin{table}[t]
  \centering
  \caption{Ablation on multi-stage pretraining for MedDINOv3 on AMOS22. }
  \label{tab:ablation_pretrain}
  \begin{tabular}{ccc|c}
    \toprule
    Stage 1: Pretraining & Stage 2: Gram Anchoring & Stage 3: Adapting to higher resolution & DSC (\%) \\
    \midrule
    $\times$ & $\times$ & $\times$  & 85.51  \\
    $\checkmark$ & $\times$ & $\times$ & 86.58 \\
    $\checkmark$ & $\checkmark$ & $\times$ & 86.54  \\
    $\checkmark$ & $\checkmark$ & $\checkmark$  & 87.38 \\
    \bottomrule
  \end{tabular}
\end{table}

\begin{figure}[h!]
    \centering
    \includegraphics[width=1.0\textwidth]{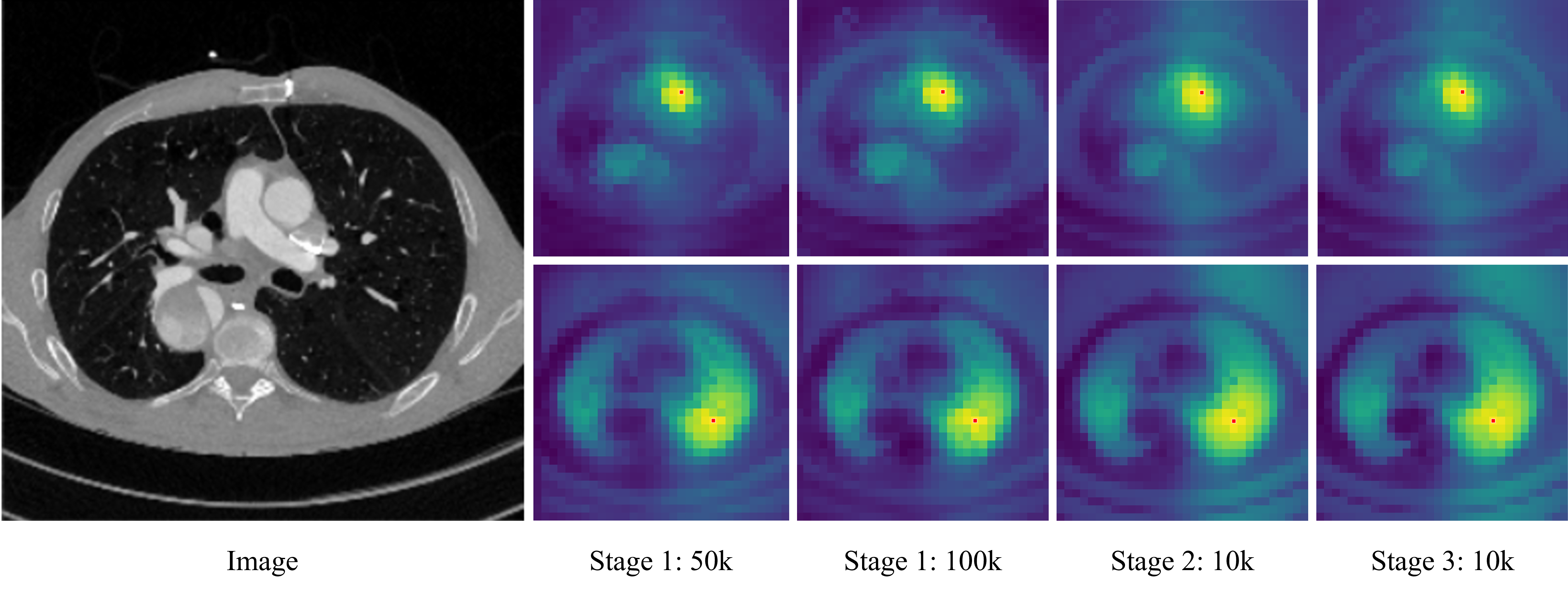}
    \caption{Evolution of the cosine similarity between the reference patch (marked in red) and all other patches. We did not observe severe patch degradation in stage 1.}
    \label{fig:ablation_patch_vis}
\end{figure}

\subsection{Ablation study}

\paragraph{Gram anchoring is optional}
We study the effect of pretraining on CT-3M on segmentation performances. As shown in Table \ref{tab:ablation_pretrain}, stage 1 pretraining improves DSC by 1.07\%, which highlights the effectiveness of DINOv2 style pretraining for learning good dense features. However, surprisingly we did not observe much gains from stage 2 with gram anchoring. We suspect that this is because the quality of patch tokens did not degrade much during stage 1 pretraining. To confirm this, we visualize them in Figure \ref{fig:ablation_patch_vis}. Nevertheless, adapting the model to higher resolution improved DSC by 0.84\% and maintained the consistency of feature maps. 


\section{Conclusion}
In this work, we present \textbf{MedDINOv3}, a simple yet effective framework for adapting vision foundation models to medical image segmentation. We refine plain Vision Transformers for medical image segmentation by proposing multi-scale token aggregation to enhance the spatial priors. We also maintain the local intricate structures by conducting high-resolution training. Building on this architecture, we curated \textbf{CT-3M}, a large-scale CT dataset, and performed domain-adaptive pretraining with DINOv3. Our systematic analysis of the three-stage pretraining recipe revealed that DINOv2-style self-distillation (Stage 1) and high-resolution adaptation (Stage 3) substantially improve feature transferability, while gram anchoring (Stage 2) provides only marginal additional benefits in our setting. Our MedDINOv3 consistently outperforms or matches strong CNN and transformer baselines in organ-at-risk segmentation, while achieving competitive performance on tumor segmentation tasks. Our results indicate that simple ViT-based architectures, when paired with domain-adaptive pretraining, can close the gap or exceed the performance of specialized CNNs. Overall, MedDINOv3 demonstrates that carefully adapting foundation models with targeted architectural refinements and domain-aligned pretraining offers a powerful and generalizable solution for medical image segmentation. 

\section*{Acknowledgments}
This research is supported in part by the National Institutes of Health under Award Numbers R01EB032680, R01DE033512, and R01CA272991.

\bibliographystyle{unsrt}  
\bibliography{references}

\end{document}